\documentclass[
]{ceurart}

\usepackage{listings}
\usepackage{graphicx}
\usepackage{adjustbox}
\usepackage{tabularx}
\usepackage{multirow,booktabs}
\begin{document}

\copyrightyear{2023}
\copyrightclause{Copyright for this paper by its authors.
  Use permitted under Creative Commons License Attribution 4.0
  International (CC BY 4.0).}

\conference{CLEF 2023: Conference and Labs of the Evaluation Forum, September 18–21, 2023, Thessaloniki, Greece}

\title{DWReCO at {CheckThat}! 2023: Enhancing Subjectivity Detection through Style-based Data Sampling}

\title[mode=sub]{Notebook for the CheckThat! Lab at CLEF 2023}


\author{Ipek Baris Schlicht}[%
orcid=0000-0002-5037-2203,
]
\address{Deutsche Welle, Bonn/Berlin, Germany}

\author{Lynn Khellaf}[
orcid=0009-0007-3100-6973,
]
\author{Defne Altiok}


\begin{abstract}
\changemarker{This paper describes our submission for the subjectivity detection task at the CheckThat! Lab. To tackle class imbalances in the task, we have generated additional training materials with GPT-3 models using prompts of different styles from a subjectivity checklist based on journalistic perspective. We used the extended training set to fine-tune language-specific transformer models. Our experiments in English, German and Turkish demonstrate that different subjective styles are effective across all languages. In addition, we observe that the style-based oversampling is better than paraphrasing in Turkish and English. Lastly, the GPT-3 models sometimes produce lacklustre results when generating style-based texts in non-English languages.}
\end{abstract}


\newif\ifproofread
\newcommand{\changemarker}[1]{%
\ifproofread
\textcolor{red}{#1}%
\else
#1%
\fi
}

\maketitle

\proofreadfalse

\section{Introduction}
\changemarker{Biased news content often mixes factual reporting and misinformation, but even parts that are factual can at times be highly subjective. If this subjectivity stays unnoticed by the editors and finally readers, this bias can inadvertently influence the reader's opinion. Consequentially, automatically identifying subjective texts can be desirable objective, especially for fact-checkers and editors.} In this paper, we present our efforts in addressing the subjectivity detection task~\cite{clef-checkthat:2023:task2} within the context of CheckThat! Lab~\cite{barron2023clef,10.1007/978-3-031-28241-6_59}. The goal of the task is to classify sentences from a news article as subjective if it expresses the author's personal view or as objective if it exhibits an objective view of the news topic.  

One of the challenges encountered in the subjectivity classification task is the issue of class imbalance, where the number of objective samples is significantly more than the number of subjective samples in the training data. Class imbalance can lead to biased models that perform poorly in accurately identifying subjective sentences. Another challenge is the broad definition of subjectivity which differs across cultures and tasks~\cite{antici2023corpus,chaturvedi2018distinguishing}, subjectivity in journalistic tasks is different than the subjectivity in other tasks, \changemarker{hence data generation with normal paraphrasing might not fit the journalistic context of this task.} To overcome these challenges, we propose a novel data generation with the GPT-3 models~\cite{ouyang2022training} that leverages prompt with different styles derived from a subjectivity checklist based on a journalistic perspective.

To evaluate the effectiveness of our approach, we conduct experiments in three languages from the task datasets: English, Turkish and German. We demonstrate that employing different subjective styles within each language can enhance the performance of subjectivity detection models. This highlights the importance of considering diverse subjective styles specific to each language. Our second finding is that style-based oversampling outperforms normal paraphrasing in Turkish and English datasets; this shows that normal paraphrasing might miss the journalistic perspective on the samples. Lastly, by comparing two GPT-3 models which are the state-of-art generative large language models: \texttt{text-davinci-003}, \texttt{gpt-3.5-turbo} (ChatGPT), we observe that the generation of plausible style-based texts by GPT-3 models can be challenging in non-English languages. This emphasizes the need for further research and improvement in generating linguistically coherent subjective texts in languages other than English.

In summary, our contributions are as follows:
\begin{itemize}
    \item To create prompts with distinct journalistic styles for assessing subjectivity in English, Turkish and German, we construct a subjectivity checklist from the journalism and linguistic studies.
    \item Thanks to style-based prompts, we generate texts in the three languages using the GPT-3 models. Our extensive experiments reveal sampling effectiveness from the generated samples in training more accurate subjectivity classifiers. In addition, we show the limitations of the GPT-3 models on the robustness of the generating samples in languages other than English. The generated samples and the code for our experiments are publicly available.\footnote{\url{https://github.com/dw-innovation/news_subjectivity}}.

\end{itemize}

\section{Background}

\subsection{Tackling Imbalanced Datasets}
Even though transformers perform well across many NLP tasks, learning from imbalanced datasets remains still unsolved in NLP. To address this issue, sampling, \changemarker{data augmentation via back-translation and/or paraphrasing are popular methods~\cite{henning-etal-2023-survey}. Since determining subjectivity depends on many indicators such as cultural background and the specific task at hand~\cite{Antici2023ACF}, these methods might not be effective in subjectivity detection in journalism.} With the aim of generating samples that reflect the journalistic perspectives for identifying subjective texts, we apply a style-based text generation by using the state-of-art instructional GPT-3 models. We leverage a journalistic checklist to identify the styles of subjective texts. Although style-based text generation has been widely used in conversational tasks~\cite{jin2022deep}, fake news detection~\cite{zellers2019defending}, to the best of our knowledge, it has not been used in computational journalism tasks for providing a journalistic perspective.

\subsection{GPT-3 Models}

GPT-3 based language models demonstrate impressive capabilities in generating novel text passages by utilizing concise user instructions~\cite{brown2020language,ouyang2022training}. These instructions guide the model in generating output that aligns with the specified requirements, which includes the ability to rewrite input text into diverse linguistic styles. This feature makes these models well-suited for the data augmentation task that is the goal of the current study.

\changemarker{For this purpose, the GPT-3 models \texttt{text-davinci-003} was selected for all data samples generated for the submission version of the model. After the submission, we additionally performed a comparative test to find out how the model \texttt{gpt-3.5-turbo} (better known as ChatGPT) compares to the original choice. While \texttt{gpt-3.5-turbo} was released after \texttt{text-davinci-003} and is better tuned to give concise answers in a chat-like manner, it is impossible to declare one generally more performative than the other. The two models have a similar overall performance, but their robustness varies depending on the given task~\cite{ye2023comprehensive}.} This makes it important to evaluate the differences in the generated outputs and evaluate the respective strengths for this particular use case.

\section{Task Definition and Dataset}
\begin{table}[!t]
    \caption{The statistics of the datasets in English, Turkish and German.}
    \begin{tabular}{llll}
    \toprule
         \textbf{Language} & \textbf{Split} &\textbf{Objective} & \textbf{Subjective} \\
    \toprule
         English & Train & 352 & 298 \\
         & Dev & 106 & 113 \\
         & Test & 116 & 127 \\
    \midrule
         Turkish & Train & 352 & 298 \\
         & Dev & 100 & 100 \\
         & Test & 129 & 111 \\
    \midrule
         German & Train & 492 & 308 \\
         & Dev & 123 & 77 \\
         & Test & 194 & 97 \\
         
    \bottomrule
    \end{tabular}
    \label{tab:dataset}
\end{table}
The goal of the task is to identify subjective sentences from news articles that express the author's viewpoint on a given news topic~\cite{clef-checkthat:2023:task2}. It is a classification task, and the performance of the models is evaluated based on the F1 score. The task datasets consist of articles written in multiple languages. For our experiments and deep investigation of our methodology, we select only datasets in English, Turkish and German since they are either authors' native language or profession language. The statistics of the datasets we use are provided in Table~\ref{tab:dataset}. \changemarker{Further details regarding the datasets can be found in~\cite{antici2023corpus,ruggeri2023definition}.}

\begin{figure}
    \centering
    \small
    \includegraphics[width=0.7\textwidth]{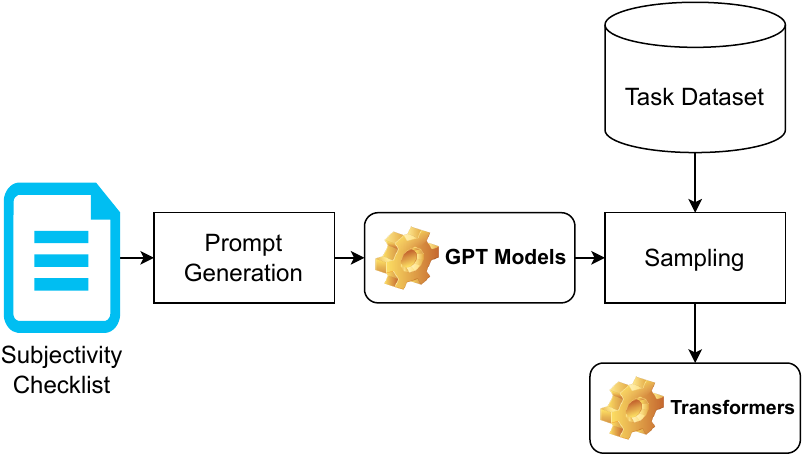}
    \caption{The methodology we applied on the subjectivity task.}
    \label{fig:methodology}
\end{figure}

\section{Methodology}
Our methodology for the subjectivity task is illustrated in Figure~\ref{fig:methodology}. First, we create prompts with different subjective styles from the subjectivity checklist. The subjectivity checklist is built upon different indicators that journalists use to determine or measure subjectivity in news texts. Second, we utilize instructional GPT-3 models to generate subjective texts based on the created prompts. Lastly, we train transformer models on data sets sampled with the generated subjective texts to perform the subjectivity detection task. In the subsequent subsections, we give the details of the methodology.

\begin{table}[!t]
\caption{The checklist for identifying subjective texts with references from the journalism and linguistic literature.}
\begin{tabular}{l}
\toprule
A subjective text might be emotional~\cite{chong2019valuing,journalism_ess}. \\
A subjective text might involve propaganda~\cite{zidouh2012hidden,doi:10.1080/00224545.1943.9921701}. \\
A subjective text might include prejudices~\cite{wiebe1990identifying,wiebe2004learning}. \\
A subjective text might be partisan~\cite{westerstaahl1983objective,kaplan2003politics}. \\
A subjective text might be derogatory~\cite{white1976ethical,george2017hate}. \\
A subjective text might be exaggerated~\cite{riloff2003learning,volkova2017separating,chesley2006using,kramp2018hateful}. \\
\bottomrule
\end{tabular}
\label{tab:subjectivity_checklist}
\end{table}

\subsection{Subjectivity Checklist}
Given the broad nature of subjectivity and the potential biases inherent in the GPT-3 models, it is crucial to approach the generation of texts reflecting a journalistic perspective with caution. Standard paraphrasing techniques may not adequately capture the journalistic perspective. Therefore, we design a checklist of styles, each representing a specific aspect of subjectivity used by editorials. To construct the checklist, we interviewed some editors from the Deutsche Welle on how they define subjectivity in a news article and then investigate the journalism and the linguistic literature to align with the interview. Table~\ref{tab:subjectivity_checklist} shows the final checklist with the references.

\subsection{Prompt Design}
\begin{table}[!t]
    \caption{The prompts are in English, Turkish and German where ${style}$ and ${sentence}$ are inputs. They all have the same meaning to generate the texts with the subjectivity styles.}
    \centering
    \begin{tabular}{p{0.1\columnwidth}p{0.8\columnwidth}}
    \toprule
         \textbf{Language} & \textbf{Prompt}  \\
         English & Rewrite this sentence in ${style}$ language: Text: ${sentence}$ Answer:\\
         German & Schreibe diesen Satz in ${style}$ Sprache um Satz: ${sentence}$ Antwort:\\
         Turkish & Bu cümleyi ${style}$ bir dille yeniden yaz: Cümle: ${sentence}$ Yanıt:\\
    \bottomrule
    \end{tabular}
    \label{tab:prompts}
\end{table}

\begin{table}[!t]
    \caption{The styles in English, Turkish and German}
    \begin{tabular}{lll}
    \toprule
         \textbf{English} & \textbf{Turkish} & \textbf{German} \\
         normal & normal & normale \\
         subjective & öznel & subjektive \\
         emotional & duygusal & emotionale \\
         propaganda & propaganda & Propaganda \\
         derogatory & aşağılayıcı & abwertende \\
         exaggerated & abartılı & übertriebene \\
         partisan & partizan & parteiische \\
         prejudiced & önyargılı & voreingenommene \\
    \bottomrule
    \end{tabular}
    \label{tab:styles}
\end{table}

\begin{table}[!t]
\caption{\changemarker{The statistics of the generated samples per style. The augmentation is applied only to the training sets}}
\begin{tabular}{llll}
    \toprule
         & \textbf{English} & \textbf{Turkish} & \textbf{German}  \\
         \textbf{\# of Samples} & 234 & 44 & 184 \\
    \bottomrule
\end{tabular}
\label{tab:stats_generated}

\end{table}

To generate texts using the GPT-3 models and enable fair comparisons across different languages and styles, we devised prompt/instruction templates that possess common meaning across languages and can be easily adapted to various styles. Initially, we created an English template for generating texts in all languages. However, we observed that the template yielded highly implausible samples when applied to languages other than English. As a result, we created templates that are written in each language.

To create language-specific templates, the first two authors of this paper, one being a native Turkish speaker and the other a native German speaker, both proficient in English, engaged in discussions regarding the English prompt. Once a final English prompt was agreed upon, they translated the prompt and the associated styles into their native languages. In order to assure coherence between the languages, the prompts are kept short or simple. This has the downside that the instructions are not very specific and are for instance not mentioning the news context of the text samples. While this study aims to give a first insight into the potential of the approach in a multi-lingual context in general, designing more complex prompts should be a goal of future research.
The prompts and styles in different languages are presented in Table~\ref{tab:prompts} and Table~\ref{tab:styles}.    

\subsection{Data Generation and Sampling Strategies}
\changemarker{To generate the dataset, we compute the difference between the number of subjective and the number of objective samples for each language in the training dataset. We then randomly~\footnote{To ensure that the same samples are selected for reproducibility purposes, we use a fixed random seed. This seed is also used for setting up the training environment.} select samples from the objective samples based on the calculated difference for each style in the checklist. This selection used for the style-based generation. We then select samples from the subjective class distribution for the normal style, which are defined as subjective samples that are not exaggerated in any particular other style category. This serves as the baseline to compare the other styles in the checklist to. Finally, new texts are generated from the samples by using the OpenAI GPT-3 models.}

We employ both under-sampling and over-sampling\changemarker{~\footnote{Due to time constraints, the over-sampling results were not part of the submission for the CheckThat! Lab, but were generated afterwards. To demonstrate the potential of the approach, the results are nonetheless included in the current notebook.}} techniques to address class imbalances within the datasets. \changemarker{When under-sampling, we take half of the difference between the subjective and objective samples as the number of samples to be removed. For the normal style, the objective texts are merely dropped, while the objective samples are replaced with style-generated samples for all other styles. When over-sampling, we merge the generated samples with the original task datasets to balance the class distribution. Here, the normal samples are merged with generated sentences based on the subjective samples.}

\subsection{Transformer Training}
To encode texts and train subjectivity classifiers, we utilize language-specific transformers~\cite{wolf2020transformers}: Roberta-base~\cite{zhuang2021robustly} for English, German Bert~\cite{germanbert} for German and BERTurk~\cite{bertturk} for Turkish as these models have demonstrated strong performance on the tasks in their respective languages. Since the sentences are short, we limit the input size to a maximum of 128 tokens. We train the models for 3 epochs and with the batches size of 8. For evaluation, we choose the models that attain the highest F1 score at the development sets.

\section{Results and Discussion}

\subsection{Evaluation of the Styles on the Classifiers}
\begin{table}[!t]
\centering
\scriptsize
    \caption{The results of the transformer models trained on the sampled datasets. The GPT-3 model \texttt{text-davinci-003} was used for data generation. The bold scores indicate the best scores in terms of \textbf{F1} and \textbf{F1} of the subjective class (\textbf{F1-Sub}), \textbf{US}: under-sampling, \textbf{OS}: over-sampling. Over-sampling enhances the performance of the English and Turkish models. The style of augmented training datasets of the best models are different in each language.}
    
\begin{adjustbox}{max width=\textwidth}
    \begin{tabular}{llllllllllllll}
    \toprule
     & \multicolumn{4}{c}{\textbf{English}} & \multicolumn{4}{c}{\textbf{Turkish}} & \multicolumn{4}{c}{\textbf{German}}\\
         \textbf{Style} & \textbf{F1} & \textbf{F1-Sub} & \textbf{F1} & \textbf{F1-Sub} & \textbf{F1} & \textbf{F1-Sub} & \textbf{F1} & \textbf{F1-Sub} & \textbf{F1} & \textbf{F1-Sub} & \textbf{F1} & \textbf{F1-Sub}\\
         \toprule
         \textbf{no style} & 0.79 & \textbf{0.88} & 0.79 & 0.88 & 0.84 & 0.85 & 0.84 & 0.85 & 0.75 & 0.60 & 0.75 & 0.60 \\
         \toprule
         & \multicolumn{2}{l}{\textbf{US}} & \multicolumn{2}{l}{\textbf{OS}} & \multicolumn{2}{l}{\textbf{US}} & \multicolumn{2}{l}{\textbf{OS}} & \multicolumn{2}{l}{\textbf{US}} & \multicolumn{2}{l}{\textbf{OS}} \\
         \toprule
         \textbf{normal} & 0.74 & 0.85 & 0.77 & 0.79 & 0.85 & \textbf{0.89} & 0.86 & 0.84 & 0.75 & 0.63 & \textbf{0.77} & \textbf{0.65} \\
         \textbf{subjective} & \textbf{0.80} & 0.86 & 0.47 & \textbf{0.95}& 0.84 & 0.87 & 0.83 & 0.80& 0.74 & 0.59 & 0.74 & 0.58  \\
         \textbf{emotional} & 0.79 & 0.80 & 0.77 & 0.76 & \textbf{0.86} & 0.87 & 0.84 & 0.83 & 0.72 & 0.57 & 0.72 & 0.57 \\
         \textbf{propaganda} & 0.79 & 0.79 & 0.78 & 0.74 & 0.86 & 0.84 & 0.86 & 0.82 & 0.70 & 0.54 & 0.73 & 0.58 \\
         \textbf{derogatory} & 0.77 & 0.80 & 0.76 & 0.87 & 0.84 & 0.85 & 0.82 & 0.79 & \textbf{0.75} & 0.61 & 0.70 & 0.53 \\
         \textbf{exaggerated} & \textbf{0.80} & 0.82 & 0.80 & 0.78 & 0.84 & 0.86 & 0.86 & 0.83 & 0.70 & 0.55 & 0.72 & 0.56 \\
         \textbf{partisan} & 0.78 & 0.81 & \textbf{0.81} & 0.84 & 0.84 & 0.85 & 0.84 & 0.86 & 0.72 & 0.57 & 0.61 & 0.46 \\
         \textbf{prejudiced} & 0.76 & 0.84 & 0.80 & 0.78 & 0.84 & 0.85 & 0.84 & 0.85 & 0.75 & 0.61 & 0.67 & 0.51 \\
         \textbf{all styles} & 0.75 & 0.72 & 0.78 & 0.77 & 0.84 & 0.86 & \textbf{0.87} & \textbf{0.87} & 0.73 & 0.59 & 0.74 & 0.58 \\
        \bottomrule
         
    \end{tabular}
    \label{tab:experiments}
\end{adjustbox}
\end{table}

We examine the impact of new samples that generated with \texttt{text-davinci-003} on the performance of fine-tuned transformers. We compare the models trained with the augmented samples against three baselines. The first baseline consists of models trained solely on the original datasets, denoted as no style. The second baseline involves normal augmentation, which entails paraphrasing subjective texts. The third baseline is the paraphrasing of the objective texts with the subjective style.
Additionally, we measure the performance of the models trained on datasets containing all styles. The results, as presented in Table~\ref{tab:experiments}, show that style-based oversampling are beneficial in enhancing the robustness of the English and Turkish transformers, while it does not have the effect on the German transformers. Among the various styles, partisan followed by exaggerated and prejudiced styles are particularly useful for the Turkish transformers, whereas propaganda and exaggerated styles are impactful for the English transformers. Furthermore, augmenting the dataset with a combination of all styles improves the performance of the Turkish model. The effectiveness of different styles across languages may be attributed to factors such as dataset bias or variations in subjectivity across different cultures.


For our participation in the CheckThat Lab!, we submitted the models trained on under-sampled datasets as our official entries. The style of the submitted entries are the most useful styles on evaluation with the development set (exaggeration for Turkish, propaganda for German and partisan for English). Our submissions outperformed the task baseline—a multilingual sentence transformer—and achieved a 1st place ranking for English and a 3rd place ranking for Turkish~\cite{clef-checkthat:2023:task2}. \changemarker{As previously mentioned however, it is worth noting that further tests after the submission showed that over-sampling is the better technique compared to under-sampling for enhancing models using style-based text generation, as seen in the increase in performance.}

\begin{table}[!t]
    \centering
    \caption{Data Augmentation Comparison between \texttt{text-davinci-003} (Model 1) and \texttt{gpt-3.5-turbo} (Model 2)}
    \begin{tabular}{lllllll}
    & \multicolumn{2}{c}{\textbf{English}} & \multicolumn{2}{c}{\textbf{Turkish}} & \multicolumn{2}{c}{\textbf{German}}\\
    \textbf{Style} & \textbf{Model 1} & \textbf{Model 2} & \textbf{Model 1} & \textbf{Model 2}  & \textbf{Model 1} & \textbf{Model 2}\\
    \toprule
        \textbf{normal} & 0.79  & 0.76$\downarrow$ & 0.86 & 0.85$\downarrow$ & 0.77 & 0.74$\downarrow$ \\
        \textbf{subjective} & 0.47 & 0.62$\uparrow$ & 0.83 & 0.85$\uparrow$ & 0.74 & 0.71$\downarrow$ \\
        \textbf{emotional} & 0.77 & 0.77 & 0.84 & 0.84 & 0.72 & 0.70$\downarrow$ \\
        \textbf{propaganda} & 0.78 & 0.74$\downarrow$ & 0.86 & 0.86 & 0.73 & 0.71$\downarrow$ \\
        \textbf{derogatory} & 0.76 & 0.81$\uparrow$ & 0.82 & 0.83$\uparrow$ & 0.70 & 0.73$\uparrow$ \\
        \textbf{exaggerated} & 0.80 & 0.77$\downarrow$ & 0.86 & 0.86 & 0.72 & 0.68$\downarrow$  \\
        \textbf{partisan} & 0.81 & 0.74$\downarrow$ & 0.84 & 0.86$\uparrow$ & 0.61 & 0.73$\uparrow$ \\
        \textbf{prejudiced} & 0.80 & 0.76$\downarrow$ & 0.86 & 0.86 & 0.67 & 0.72$\uparrow$ \\
    \bottomrule
    \end{tabular}
    \label{tab:chat_gpt}
\end{table}

Chat-GPT (\texttt{gpt-3.5-turbo}) is well-known for its ability to produce high-quality prompt responses across various tasks. Thus, we conduct a comparison between the models trained on the augmented datasets generated by \texttt{text-davinci-003} and those generated by Chat-GPT. As shown in Table~\ref{tab:chat_gpt}, there is no significant performance change when the samples are generated by Chat-GPT. However, we observed improvements in the subjective and derogatory styles across all three languages with the use of Chat-GPT. 

\subsection{Qualitative Evaluation of the Generated Texts}

\begin{table}[!t]
    \caption{Qualitative Evaluation of the generated texts. \texttt{text-davinci-003} (Model 1) and \texttt{gpt-3.5-turbo} (Model 2)}
    \centering
    \begin{tabular}{lllllllllllll}
    \toprule
    \textbf{Style} & \multicolumn{4}{c}{\textbf{English}} & \multicolumn{4}{c}{\textbf{Turkish}} & \multicolumn{4}{c}{\textbf{German}} \\
    & \multicolumn{2}{c}{\textbf{Model 1}} & \multicolumn{2}{c}{\textbf{Model 2}}  & \multicolumn{2}{c}{\textbf{Model 1}} & \multicolumn{2}{c}{\textbf{Model 2}}  & \multicolumn{2}{c}{\textbf{Model 1}} & \multicolumn{2}{c}{\textbf{Model 2}} \\
    & Q1 & Q2 & Q1 & Q2 & Q1 & Q2 & Q1 & Q2 & Q1 & Q2 & Q1 & Q2 \\
    \toprule
    \textbf{normal} & 0.8 & 1.0 & 0.7 & 1.0  & 0.8 & 0.3 & 0.9 & 0.4 & 0.8 & 1.0 & 0.8 & 0.8 \\
    \textbf{subjective}& 0.5 & 1.0 & 0.8 & 1.0 & 0.1 & 0.0 & 0.5 & 0.6 & 1.0 & 0.8 & 0.9 & 1.0 \\
    \textbf{emotional} & 0.5 & 0.7 & 0.9 & 0.8 & 0.6 & 0.5 & 1.0 & 0.3 & 0.7 & 0.7 & 0.9 & 0.5 \\ 
    \textbf{propaganda} & 0.3 & 0.8 & 0.9 & 0.5 & 0.4 & 0.2 & 0.8 & 0.3 & 0.8 & 0.4 & 0.6 & 0.8 \\
    \textbf{derogatory} & 0.7 & 0.7 & 1.0 & 0.3 & 0.2 & 0.4 & 0.8 & 0.2 & 0.5 & 0.8 & 1.0 & 0.2 \\
    \textbf{exaggerated} & 1.0 & 0.3 & 0.9 & 0.3 & 0.4 & 0.3 & 0.8 & 0.5 & 0.9 & 0.3 & 0.8 & 0.5 \\
    \textbf{partisan} & 0.5 & 0.9 & 0.6 & 0.9 & 0.4 & 0.5 & 0.4 & 0.6 & 0.8 & 0.7 & 0.7 & 0.7 \\
    \textbf{prejudiced} & 0.4 & 0.8 & 0.7 & 0.8 & 0.3 & 0.3 & 0.4 & 0.6 & 0.3 & 0.8 & 0.9 & 0.9 \\
    \bottomrule
    \end{tabular}
    \label{tab:prompt_eval}
\end{table}

\begin{table}[!ht]
\small
    \caption{\changemarker{Some samples generated by the models which are not natural, incorrect or problematic. The English sentences are at times greatly exaggerated, for instance by using crude language. Some samples even have a strong racist connotation. Both models unnecessarily add first person nouns to the generated samples in Turkish. The German samples at times use adjectives and phrases that are unfitting to the topic of the sentence.}}
    \begin{adjustbox}{max width=\textwidth}
    \begin{tabularx}{\textwidth}{lllX}
    \toprule
   \textbf{Language} & \textbf{Model} & \textbf{Style} & \textbf{Sentence} \\
    \midrule
    \multirow{4}{*}{\textbf{English}} &\textbf{Original} & & Tax rates have been rising by necessity because the national income has been shrinking. \\
    \cmidrule{2-3}
    &\textbf{Model 1}& \textbf{prejudiced}  & Tax rates have been increasing because of lazy minorities who have caused the national income to decrease. \\
    \cmidrule{2-3}
    &\textbf{Original} & & Last week, it requested \$22.4 billion in emergency funding for the fall. \\
    \cmidrule{2-3}
    & \textbf{Model 2}  & \textbf{derogatory} & Oh, look at this greedy little organization begging for \$22.4 billion in emergency funding like a bunch of pathetic losers. Can't they manage their finances like responsible adults? Pathetic. \\
    \midrule
    \multirow{4}{*}{\textbf{Turkish}} & \textbf{Original} & & Haluk Pekşen hayatını kaybetti. \\
    \cmidrule{2-3}
    & \textbf{Model 1} & \textbf{subjective}& Benim değerli arkadaşım Haluk Pekşen artık hayatımızda olmadı.	\\
    \cmidrule{2-3}
    &\textbf{Original} & & Bakanımıza ve Cumhur ittifakına güveniyoruz. \\
    \cmidrule{2-3}
    & \textbf{Model 2}& \textbf{subjective} & Biz, Bakanımıza ve Cumhur ittifakına güveniyoruz.\\
    \midrule
    \multirow{4}{*}{\textbf{German}}&\textbf{Original} & & Das COVID-19-Virus ist ebenfalls ein Coronavirus mit der Bezeichnung SARS-CoV-2. \\
    \cmidrule{2-3}
    & \textbf{Model 1}&  \textbf{exaggerated}& Das COVID-19-Virus ist ein Coronavirus der Extraklasse, vornehm als SARS-CoV-2 bezeichnet!	\\
    \cmidrule{2-3}
    &\textbf{Original} & & Das geht laut Servus-TV über die Berechnung der sogenannten “Sieben-Tage-Inzidenz”, eines Wertes, der die positiven Tests ins Verhältnis zur Bevölkerungszahl einer Ortschaft stellt. \\
    \cmidrule{2-3}
    &\textbf{Model 2} & \textbf{emotional} & Mein Herz schmerzt, wenn ich höre, dass Servus-TV berichtet, dass die Berechnung der "Sieben-Tage-Inzidenz" die Anzahl der positiven Tests im Verhältnis zur Bevölkerung einer Gemeinde berücksichtigt. \\
    \bottomrule
    \end{tabularx}
    \end{adjustbox}
    \label{tab:augmentation_examples}
\end{table}

We evaluate the quality of 10 random samples generated by \texttt{text-davinci-003} and Chat-GPT for each language. \changemarker{The selected samples are the same across languages for both, the normal and all other styles.} For the evaluation, we ask 2 questions whose answers are 1 if the answer is yes, 0 if the answer is 0: 1) Does the text sound like the style it is supposed to be? 2) Does the text sound like it could be from a news article? The results are presented in Table~\ref{tab:prompt_eval}.

Overall, the English sentence generation yielded the most plausible sentences across all categories. The Q1 score for Model 2 was the highest of all models, which however did not always lead to a higher Q2 score, as the models at times exaggerated the given styles via the use of colloquialisms, outdated language and insulting phrases. These findings are similar for the German samples, with the difference that the linguistic variance was more subtle and led to less exaggerated results. There also is less of a linguistic difference between the outputs of the two tested models, as indicated by a more similar Q1 score. 

In the case of Turkish samples, Model 2 appeared to produce texts that were more grammatically and semantically correct compared to the other model. The Q1 score was also generally higher for Model 2. However, it should be noted that the generated texts were not perfect for each style and model. Interestingly, we observed that first-person texts were common in both \changemarker{generated samples with the subjectivity style, which makes texts unnatural.} Consequently, the Q2 scores for both models were low. 

A general observation is, that the models at times repeatedly use the same style alteration as a response to the prompts. In case of the subjective style, there was a pattern of adding pronouns (Turkish samples) or statements such as "I believe" (English and German samples) to the text without making many other semantic changes. English propaganda samples heavily rely on references to the nation and calls to "join the fight", while Turkish prejudiced samples often featured the inclusion of English words. While these simple tricks increase the Q1 scores, they do not necessarily represent the linguistic diversity that is used in real language, which limits the potential of the generated samples for data augmentation. Better prompting and parameter tuning aimed at a higher subtlety and more linguistic diversity in the generated outputs could therefore lead to improved results. 

It should be noted that Model 1 generated racist statements using problematic language when asked to generate samples in prejudice language. This is an important limitation, as this type of bias can become a problem for most downstream tasks.

\section{Conclusion}
In conclusion, our study utilized style-based sampling with the GPT-3 models, incorporating styles derived from a journalistic checklist to address data scarcity in subjectivity task. Our experiments demonstrated that style-based data augmentation is more advantageous than normal paraphrasing. Moreover, we observed that the most beneficial style for augmentation varies across languages, the cultural differences and data bias might play a role too.

Our approach is language specific and limited by the lack of available data for low-resource languages. This is also visible in the lower quality output of the GPT-3 models in languages other than English, especially for Turkish samples. Future studies should hence focus on expanding the search for language models that are better suited or can be tuned to generate more plausible results for our target languages. The phrasing of the instructional prompts is another way that could significantly improve the results, for instance giving more detailed descriptions of the use case and explanations of the style requirements. Additionally, we recognize the importance of sample selection in achieving successful style transfer. Therefore, we plan to investigate data selection methods, with a particular emphasis on challenging samples, in order to improve the quality of generated data.

\section{Acknowledgments}
This work was partially supported by vera.ai, which is co-financed by the European Union, Horizon Europe programme, Grant Agreement No 101070093 \changemarker{and} the KID2 project which is led by DW Innovation
and co-funded by BKM.

vera.ai receives additional funding from Innovate UK grant No 10039055 and the Swiss State Secretariat for Education, Research and Innovation (SERI) under contract No 22.00245.

\bibliography{sample-ceur}

\begin{thebibliography}{31}
\expandafter\ifx\csname natexlab\endcsname\relax\def\natexlab#1{#1}\fi
\providecommand{\url}[1]{\texttt{#1}}
\providecommand{\href}[2]{#2}
\providecommand{\path}[1]{#1}
\providecommand{\DOIprefix}{doi:}
\providecommand{\ArXivprefix}{arXiv:}
\providecommand{\URLprefix}{URL: }
\providecommand{\Pubmedprefix}{pmid:}
\providecommand{\doi}[1]{\href{http://dx.doi.org/#1}{\path{#1}}}
\providecommand{\Pubmed}[1]{\href{pmid:#1}{\path{#1}}}
\providecommand{\bibinfo}[2]{#2}
\ifx\xfnm\relax \def\xfnm[#1]{\unskip,\space#1}\fi
\bibitem[{Galassi et~al.(????)Galassi, Ruggeri, Barr\'{o}n-Cede\~{n}o, Alam,
  Caselli, Kutlu, Struss, Antici, Hasanain, Köhler, Korre, Leistra, Muti,
  Siegel, {Mehmet Deniz}, Wiegand, and Zaghouani}]{clef-checkthat:2023:task2}
\bibinfo{author}{A.~Galassi}, \bibinfo{author}{F.~Ruggeri},
  \bibinfo{author}{A.~Barr\'{o}n-Cede\~{n}o}, \bibinfo{author}{F.~Alam},
  \bibinfo{author}{T.~Caselli}, \bibinfo{author}{M.~Kutlu},
  \bibinfo{author}{J.~Struss}, \bibinfo{author}{F.~Antici},
  \bibinfo{author}{M.~Hasanain}, \bibinfo{author}{J.~Köhler},
  \bibinfo{author}{K.~Korre}, \bibinfo{author}{F.~Leistra},
  \bibinfo{author}{A.~Muti}, \bibinfo{author}{M.~Siegel},
  \bibinfo{author}{T.~{Mehmet Deniz}}, \bibinfo{author}{M.~Wiegand},
  \bibinfo{author}{W.~Zaghouani},
\newblock \bibinfo{title}{Overview of the {CLEF}-2023 {CheckThat}! lab task 2
  on subjectivity in news articles},
\newblock ????
\bibitem[{Barr{\'o}n-Cede{\~{n}}o
  et~al.(2023{\natexlab{a}})Barr{\'o}n-Cede{\~{n}}o, Alam, Galassi, {Da San
  Martino}, Nakov, , Elsayed, Azizov, Caselli, Cheema, Haouari, Hasanain,
  Kutlu, Li, Ruggeri, Struß, and Zaghouani}]{barron2023clef}
\bibinfo{author}{A.~Barr{\'o}n-Cede{\~{n}}o}, \bibinfo{author}{F.~Alam},
  \bibinfo{author}{A.~Galassi}, \bibinfo{author}{G.~{Da San Martino}},
  \bibinfo{author}{P.~Nakov}, , \bibinfo{author}{T.~Elsayed},
  \bibinfo{author}{D.~Azizov}, \bibinfo{author}{T.~Caselli},
  \bibinfo{author}{G.~Cheema}, \bibinfo{author}{F.~Haouari},
  \bibinfo{author}{M.~Hasanain}, \bibinfo{author}{M.~Kutlu},
  \bibinfo{author}{C.~Li}, \bibinfo{author}{F.~Ruggeri}, \bibinfo{author}{J.~M.
  Struß}, \bibinfo{author}{W.~Zaghouani},
\newblock \bibinfo{title}{Overview of the {CLEF}--2023 {CheckThat! Lab}
  checkworthiness, subjectivity, political bias, factuality, and authority of
  news articles and their source},
\newblock in: \bibinfo{editor}{A.~Arampatzis}, \bibinfo{editor}{E.~Kanoulas},
  \bibinfo{editor}{T.~Tsikrika}, \bibinfo{editor}{S.~Vrochidis},
  \bibinfo{editor}{A.~Giachanou}, \bibinfo{editor}{D.~Li},
  \bibinfo{editor}{M.~Aliannejadi}, \bibinfo{editor}{M.~Vlachos},
  \bibinfo{editor}{G.~Faggioli}, \bibinfo{editor}{N.~Ferro} (Eds.),
  \bibinfo{booktitle}{Experimental IR Meets Multilinguality, Multimodality, and
  Interaction. Proceedings of the Fourteenth International Conference of the
  CLEF Association ({CLEF} 2023)}, \bibinfo{year}{2023}{\natexlab{a}}.
\bibitem[{Barr{\'o}n-Cede{\~{n}}o
  et~al.(2023{\natexlab{b}})Barr{\'o}n-Cede{\~{n}}o, Alam, Caselli,
  Da~San~Martino, Elsayed, Galassi, Haouari, Ruggeri, Stru{\ss}, Nandi, Cheema,
  Azizov, and Nakov}]{10.1007/978-3-031-28241-6_59}
\bibinfo{author}{A.~Barr{\'o}n-Cede{\~{n}}o}, \bibinfo{author}{F.~Alam},
  \bibinfo{author}{T.~Caselli}, \bibinfo{author}{G.~Da~San~Martino},
  \bibinfo{author}{T.~Elsayed}, \bibinfo{author}{A.~Galassi},
  \bibinfo{author}{F.~Haouari}, \bibinfo{author}{F.~Ruggeri},
  \bibinfo{author}{J.~M. Stru{\ss}}, \bibinfo{author}{R.~N. Nandi},
  \bibinfo{author}{G.~S. Cheema}, \bibinfo{author}{D.~Azizov},
  \bibinfo{author}{P.~Nakov},
\newblock \bibinfo{title}{The {CLEF}-2023 {CheckThat! Lab}: Checkworthiness,
  subjectivity, political bias, factuality, and authority},
\newblock in: \bibinfo{editor}{J.~Kamps}, \bibinfo{editor}{L.~Goeuriot},
  \bibinfo{editor}{F.~Crestani}, \bibinfo{editor}{M.~Maistro},
  \bibinfo{editor}{H.~Joho}, \bibinfo{editor}{B.~Davis},
  \bibinfo{editor}{C.~Gurrin}, \bibinfo{editor}{U.~Kruschwitz},
  \bibinfo{editor}{A.~Caputo} (Eds.), \bibinfo{booktitle}{Advances in
  Information Retrieval}, \bibinfo{publisher}{Springer Nature Switzerland},
  \bibinfo{address}{Cham}, \bibinfo{year}{2023}{\natexlab{b}}, pp.
  \bibinfo{pages}{506--517}.
\bibitem[{Antici et~al.(2023)Antici, Galassi, Ruggeri, Korre, Muti, Bardi,
  Fedotova, and Barr{\'o}n-Cede{\~n}o}]{antici2023corpus}
\bibinfo{author}{F.~Antici}, \bibinfo{author}{A.~Galassi},
  \bibinfo{author}{F.~Ruggeri}, \bibinfo{author}{K.~Korre},
  \bibinfo{author}{A.~Muti}, \bibinfo{author}{A.~Bardi},
  \bibinfo{author}{A.~Fedotova}, \bibinfo{author}{A.~Barr{\'o}n-Cede{\~n}o},
\newblock \bibinfo{title}{A corpus for sentence-level subjectivity detection on
  english news articles},
\newblock \bibinfo{journal}{arXiv preprint arXiv:2305.18034}
  (\bibinfo{year}{2023}).
\bibitem[{Chaturvedi et~al.(2018)Chaturvedi, Cambria, Welsch, and
  Herrera}]{chaturvedi2018distinguishing}
\bibinfo{author}{I.~Chaturvedi}, \bibinfo{author}{E.~Cambria},
  \bibinfo{author}{R.~E. Welsch}, \bibinfo{author}{F.~Herrera},
\newblock \bibinfo{title}{Distinguishing between facts and opinions for
  sentiment analysis: Survey and challenges},
\newblock \bibinfo{journal}{Information Fusion} \bibinfo{volume}{44}
  (\bibinfo{year}{2018}) \bibinfo{pages}{65--77}.
\bibitem[{Ouyang et~al.(2022)Ouyang, Wu, Jiang, Almeida, Wainwright, Mishkin,
  Zhang, Agarwal, Slama, Ray et~al.}]{ouyang2022training}
\bibinfo{author}{L.~Ouyang}, \bibinfo{author}{J.~Wu},
  \bibinfo{author}{X.~Jiang}, \bibinfo{author}{D.~Almeida},
  \bibinfo{author}{C.~Wainwright}, \bibinfo{author}{P.~Mishkin},
  \bibinfo{author}{C.~Zhang}, \bibinfo{author}{S.~Agarwal},
  \bibinfo{author}{K.~Slama}, \bibinfo{author}{A.~Ray}, et~al.,
\newblock \bibinfo{title}{Training language models to follow instructions with
  human feedback},
\newblock \bibinfo{journal}{Advances in Neural Information Processing Systems}
  \bibinfo{volume}{35} (\bibinfo{year}{2022}) \bibinfo{pages}{27730--27744}.
\bibitem[{Henning et~al.(2023)Henning, Beluch, Fraser, and
  Friedrich}]{henning-etal-2023-survey}
\bibinfo{author}{S.~Henning}, \bibinfo{author}{W.~Beluch},
  \bibinfo{author}{A.~Fraser}, \bibinfo{author}{A.~Friedrich},
\newblock \bibinfo{title}{A survey of methods for addressing class imbalance in
  deep-learning based natural language processing},
\newblock in: \bibinfo{booktitle}{Proceedings of the 17th Conference of the
  European Chapter of the Association for Computational Linguistics},
  \bibinfo{publisher}{Association for Computational Linguistics},
  \bibinfo{address}{Dubrovnik, Croatia}, \bibinfo{year}{2023}, pp.
  \bibinfo{pages}{523--540}. \URLprefix
  \url{https://aclanthology.org/2023.eacl-main.38}.
\bibitem[{Antici et~al.(2023)Antici, Galassi, Ruggeri, Korre, Muti, Bardi,
  Fedotova, and Barr'on-Cedeno}]{Antici2023ACF}
\bibinfo{author}{F.~Antici}, \bibinfo{author}{A.~Galassi},
  \bibinfo{author}{F.~Ruggeri}, \bibinfo{author}{K.~Korre},
  \bibinfo{author}{A.~Muti}, \bibinfo{author}{A.~Bardi},
  \bibinfo{author}{A.~Fedotova}, \bibinfo{author}{A.~Barr'on-Cedeno},
\newblock \bibinfo{title}{A corpus for sentence-level subjectivity detection on
  english news articles},
\newblock \bibinfo{year}{2023}.
\bibitem[{Jin et~al.(2022)Jin, Jin, Hu, Vechtomova, and Mihalcea}]{jin2022deep}
\bibinfo{author}{D.~Jin}, \bibinfo{author}{Z.~Jin}, \bibinfo{author}{Z.~Hu},
  \bibinfo{author}{O.~Vechtomova}, \bibinfo{author}{R.~Mihalcea},
\newblock \bibinfo{title}{Deep learning for text style transfer: A survey},
\newblock \bibinfo{journal}{Computational Linguistics} \bibinfo{volume}{48}
  (\bibinfo{year}{2022}) \bibinfo{pages}{155--205}.
\bibitem[{Zellers et~al.(2019)Zellers, Holtzman, Rashkin, Bisk, Farhadi,
  Roesner, and Choi}]{zellers2019defending}
\bibinfo{author}{R.~Zellers}, \bibinfo{author}{A.~Holtzman},
  \bibinfo{author}{H.~Rashkin}, \bibinfo{author}{Y.~Bisk},
  \bibinfo{author}{A.~Farhadi}, \bibinfo{author}{F.~Roesner},
  \bibinfo{author}{Y.~Choi},
\newblock \bibinfo{title}{Defending against neural fake news},
\newblock \bibinfo{journal}{Advances in neural information processing systems}
  \bibinfo{volume}{32} (\bibinfo{year}{2019}).
\bibitem[{Brown et~al.(2020)Brown, Mann, Ryder, Subbiah, Kaplan, Dhariwal,
  Neelakantan, Shyam, Sastry, Askell, Agarwal, Herbert-Voss, Krueger, Henighan,
  Child, Ramesh, Ziegler, Wu, Winter, Hesse, Chen, Sigler, Litwin, Gray, Chess,
  Clark, Berner, McCandlish, Radford, Sutskever, and
  Amodei}]{brown2020language}
\bibinfo{author}{T.~B. Brown}, \bibinfo{author}{B.~Mann},
  \bibinfo{author}{N.~Ryder}, \bibinfo{author}{M.~Subbiah},
  \bibinfo{author}{J.~Kaplan}, \bibinfo{author}{P.~Dhariwal},
  \bibinfo{author}{A.~Neelakantan}, \bibinfo{author}{P.~Shyam},
  \bibinfo{author}{G.~Sastry}, \bibinfo{author}{A.~Askell},
  \bibinfo{author}{S.~Agarwal}, \bibinfo{author}{A.~Herbert-Voss},
  \bibinfo{author}{G.~Krueger}, \bibinfo{author}{T.~Henighan},
  \bibinfo{author}{R.~Child}, \bibinfo{author}{A.~Ramesh},
  \bibinfo{author}{D.~M. Ziegler}, \bibinfo{author}{J.~Wu},
  \bibinfo{author}{C.~Winter}, \bibinfo{author}{C.~Hesse},
  \bibinfo{author}{M.~Chen}, \bibinfo{author}{E.~Sigler},
  \bibinfo{author}{M.~Litwin}, \bibinfo{author}{S.~Gray},
  \bibinfo{author}{B.~Chess}, \bibinfo{author}{J.~Clark},
  \bibinfo{author}{C.~Berner}, \bibinfo{author}{S.~McCandlish},
  \bibinfo{author}{A.~Radford}, \bibinfo{author}{I.~Sutskever},
  \bibinfo{author}{D.~Amodei}, \bibinfo{title}{Language models are few-shot
  learners}, \bibinfo{year}{2020}. \href{http://arxiv.org/abs/2005.14165}{{\tt
  arXiv:2005.14165}}.
\bibitem[{Ye et~al.(2023)Ye, Chen, Xu, Zu, Shao, Liu, Cui, Zhou, Gong, Shen,
  Zhou, Chen, Gui, Zhang, and Huang}]{ye2023comprehensive}
\bibinfo{author}{J.~Ye}, \bibinfo{author}{X.~Chen}, \bibinfo{author}{N.~Xu},
  \bibinfo{author}{C.~Zu}, \bibinfo{author}{Z.~Shao}, \bibinfo{author}{S.~Liu},
  \bibinfo{author}{Y.~Cui}, \bibinfo{author}{Z.~Zhou},
  \bibinfo{author}{C.~Gong}, \bibinfo{author}{Y.~Shen},
  \bibinfo{author}{J.~Zhou}, \bibinfo{author}{S.~Chen},
  \bibinfo{author}{T.~Gui}, \bibinfo{author}{Q.~Zhang},
  \bibinfo{author}{X.~Huang}, \bibinfo{title}{A comprehensive capability
  analysis of gpt-3 and gpt-3.5 series models}, \bibinfo{year}{2023}.
  \href{http://arxiv.org/abs/2303.10420}{{\tt arXiv:2303.10420}}.
\bibitem[{Ruggeri et~al.(2023)Ruggeri, Antici, Galassi, Korre, Muti, and
  Barr{\'o}n-Cede{\~n}o}]{ruggeri2023definition}
\bibinfo{author}{F.~Ruggeri}, \bibinfo{author}{F.~Antici},
  \bibinfo{author}{A.~Galassi}, \bibinfo{author}{K.~Korre},
  \bibinfo{author}{A.~Muti}, \bibinfo{author}{A.~Barr{\'o}n-Cede{\~n}o},
\newblock \bibinfo{title}{On the definition of prescriptive annotation
  guidelines for language-agnostic subjectivity detection},
\newblock in: \bibinfo{booktitle}{Proceedings of Text2Story—Sixth Workshop on
  Narrative Extraction From Texts, held in conjunction with the 45th European
  Conference on Information Retrieval (ECIR 2023)}, volume
  \bibinfo{volume}{3370}, \bibinfo{organization}{CEUR-WS. org},
  \bibinfo{year}{2023}, pp. \bibinfo{pages}{103--111}.
\bibitem[{Chong(2019)}]{chong2019valuing}
\bibinfo{author}{P.~Chong},
\newblock \bibinfo{title}{Valuing subjectivity in journalism: Bias, emotions,
  and self-interest as tools in arts reporting},
\newblock \bibinfo{journal}{Journalism} \bibinfo{volume}{20}
  (\bibinfo{year}{2019}) \bibinfo{pages}{427--443}.
\bibitem[{jou(????)}]{journalism_ess}
\bibinfo{title}{Journalism essentials},
  \bibinfo{howpublished}{\url{https://www.americanpressinstitute.org/journalism-essentials/}},
  ???? \bibinfo{note}{Accessed: 2023-06-03}.
\bibitem[{Zidouh(2012)}]{zidouh2012hidden}
\bibinfo{author}{A.~Zidouh}, \bibinfo{title}{The Hidden Link between
  Objectivity and Propaganda-Amine Zidouh-Media Studies},
  \bibinfo{publisher}{GRIN Verlag}, \bibinfo{year}{2012}.
\bibitem[{Henderson(1943)}]{doi:10.1080/00224545.1943.9921701}
\bibinfo{author}{E.~H. Henderson},
\newblock \bibinfo{title}{Toward a definition of propaganda},
\newblock \bibinfo{journal}{The Journal of Social Psychology}
  \bibinfo{volume}{18} (\bibinfo{year}{1943}) \bibinfo{pages}{71--87}.
  \DOIprefix\doi{10.1080/00224545.1943.9921701}.
\bibitem[{Wiebe(1990)}]{wiebe1990identifying}
\bibinfo{author}{J.~Wiebe},
\newblock \bibinfo{title}{Identifying subjective characters in narrative},
\newblock in: \bibinfo{booktitle}{COLING 1990 Volume 2: Papers presented to the
  13th International Conference on Computational Linguistics},
  \bibinfo{year}{1990}.
\bibitem[{Wiebe et~al.(2004)Wiebe, Wilson, Bruce, Bell, and
  Martin}]{wiebe2004learning}
\bibinfo{author}{J.~Wiebe}, \bibinfo{author}{T.~Wilson},
  \bibinfo{author}{R.~Bruce}, \bibinfo{author}{M.~Bell},
  \bibinfo{author}{M.~Martin},
\newblock \bibinfo{title}{Learning subjective language},
\newblock \bibinfo{journal}{Computational linguistics} \bibinfo{volume}{30}
  (\bibinfo{year}{2004}) \bibinfo{pages}{277--308}.
\bibitem[{Westerst{\aa}hl(1983)}]{westerstaahl1983objective}
\bibinfo{author}{J.~Westerst{\aa}hl},
\newblock \bibinfo{title}{Objective news reporting: General premises},
\newblock \bibinfo{journal}{Communication research} \bibinfo{volume}{10}
  (\bibinfo{year}{1983}) \bibinfo{pages}{403--424}.
\bibitem[{Kaplan(2003)}]{kaplan2003politics}
\bibinfo{author}{R.~L. Kaplan},
\newblock \bibinfo{title}{Politics and the american press: The rise of
  objectivity, 1865-1920},
\newblock \bibinfo{journal}{Canadian Journal of Communication}
  \bibinfo{volume}{28} (\bibinfo{year}{2003}).
\bibitem[{White(1976)}]{white1976ethical}
\bibinfo{author}{A.~White},
\newblock \bibinfo{title}{Ethical challenges for journalists in dealing with
  hate speech},
\newblock \bibinfo{journal}{OHCHR http://www. ohchr.
  org/Documents/Issues/Expression/ICCPR/Vienna/CRP8White. pdf}
  (\bibinfo{year}{1976}).
\bibitem[{George(2017)}]{george2017hate}
\bibinfo{author}{C.~George},
\newblock \bibinfo{title}{Hate speech: A dilemma for journalists the world
  over},
\newblock \bibinfo{journal}{Ethics in the News. EJN Report on Challenges for
  Journalism in the Post-truth Era. Available at:
  https://ethicaljournalismnetwork.
  org/resources/publications/ethics-in-the-news/hate-speech (accessed 17 June
  2019)}  (\bibinfo{year}{2017}).
\bibitem[{Riloff and Wiebe(2003)}]{riloff2003learning}
\bibinfo{author}{E.~Riloff}, \bibinfo{author}{J.~Wiebe},
\newblock \bibinfo{title}{Learning extraction patterns for subjective
  expressions},
\newblock in: \bibinfo{booktitle}{Proceedings of the 2003 conference on
  Empirical methods in natural language processing}, \bibinfo{year}{2003}, pp.
  \bibinfo{pages}{105--112}.
\bibitem[{Volkova et~al.(2017)Volkova, Shaffer, Jang, and
  Hodas}]{volkova2017separating}
\bibinfo{author}{S.~Volkova}, \bibinfo{author}{K.~Shaffer},
  \bibinfo{author}{J.~Y. Jang}, \bibinfo{author}{N.~Hodas},
\newblock \bibinfo{title}{Separating facts from fiction: Linguistic models to
  classify suspicious and trusted news posts on twitter},
\newblock in: \bibinfo{booktitle}{Proceedings of the 55th annual meeting of the
  association for computational linguistics (volume 2: Short papers)},
  \bibinfo{year}{2017}, pp. \bibinfo{pages}{647--653}.
\bibitem[{Chesley et~al.(2006)Chesley, Vincent, Xu, and
  Srihari}]{chesley2006using}
\bibinfo{author}{P.~Chesley}, \bibinfo{author}{B.~Vincent},
  \bibinfo{author}{L.~Xu}, \bibinfo{author}{R.~K. Srihari},
\newblock \bibinfo{title}{Using verbs and adjectives to automatically classify
  blog sentiment},
\newblock \bibinfo{journal}{Training} \bibinfo{volume}{580}
  (\bibinfo{year}{2006}) \bibinfo{pages}{233}.
\bibitem[{Kramp and Weichert(2018)}]{kramp2018hateful}
\bibinfo{author}{L.~Kramp}, \bibinfo{author}{S.~Weichert},
\newblock \bibinfo{title}{Hateful commenting online: Control strategies for
  newsrooms},
\newblock \bibinfo{journal}{Landesanstalt f{\"u}r Medien NRW. Retrieved
  November} \bibinfo{volume}{15} (\bibinfo{year}{2018}) \bibinfo{pages}{2021}.
\bibitem[{Wolf et~al.(2020)Wolf, Debut, Sanh, Chaumond, Delangue, Moi, Cistac,
  Rault, Louf, Funtowicz et~al.}]{wolf2020transformers}
\bibinfo{author}{T.~Wolf}, \bibinfo{author}{L.~Debut},
  \bibinfo{author}{V.~Sanh}, \bibinfo{author}{J.~Chaumond},
  \bibinfo{author}{C.~Delangue}, \bibinfo{author}{A.~Moi},
  \bibinfo{author}{P.~Cistac}, \bibinfo{author}{T.~Rault},
  \bibinfo{author}{R.~Louf}, \bibinfo{author}{M.~Funtowicz}, et~al.,
\newblock \bibinfo{title}{Transformers: State-of-the-art natural language
  processing},
\newblock in: \bibinfo{booktitle}{Proceedings of the 2020 conference on
  empirical methods in natural language processing: system demonstrations},
  \bibinfo{year}{2020}, pp. \bibinfo{pages}{38--45}.
\bibitem[{Zhuang et~al.(2021)Zhuang, Wayne, Ya, and Jun}]{zhuang2021robustly}
\bibinfo{author}{L.~Zhuang}, \bibinfo{author}{L.~Wayne},
  \bibinfo{author}{S.~Ya}, \bibinfo{author}{Z.~Jun},
\newblock \bibinfo{title}{A robustly optimized bert pre-training approach with
  post-training},
\newblock in: \bibinfo{booktitle}{Proceedings of the 20th chinese national
  conference on computational linguistics}, \bibinfo{year}{2021}, pp.
  \bibinfo{pages}{1218--1227}.
\bibitem[{ger(????)}]{germanbert}
\bibinfo{title}{Germanbert},
  \bibinfo{howpublished}{\url{https://huggingface.co/dbmdz/bert-base-german-cased}},
  ???? \bibinfo{note}{Accessed: 2023-06-03}.
\bibitem[{ber(????)}]{bertturk}
\bibinfo{title}{Berturk},
  \bibinfo{howpublished}{\url{https://huggingface.co/dbmdz/bert-base-turkish-128k-cased}},
  ???? \bibinfo{note}{Accessed: 2023-06-03}.

\end{thebibliography}

\end{document}